\pdfoutput=1

\documentclass[11pt]{article}

\usepackage[]{acl}

\usepackage{times}
\usepackage{latexsym}
\usepackage{booktabs}
\usepackage[T1]{fontenc}

\usepackage[utf8]{inputenc}

\usepackage{microtype}
\usepackage{graphicx}
\usepackage{amsfonts}
\usepackage{amsmath}
\usepackage{amssymb}
\usepackage[ruled,linesnumbered]{algorithm2e}
\usepackage{caption}
\usepackage{graphicx}
\usepackage{float}
\usepackage{multirow}
\usepackage{graphicx,subfigure}

\usepackage{graphicx}

\title{Fine-Tuning Language Models with Reward Learning on Policy}

\author{
Hao Lang \quad
Fei Huang \quad
Yongbin Li \thanks{\quad Corresponding author.} \\
Alibaba Group \\
\texttt{\{hao.lang, f.huang, shuide.lyb\}@alibaba-inc.com}
}

\begin{document}
\maketitle
\begin{abstract}

Reinforcement learning from human feedback (RLHF) has emerged as an effective approach to aligning large language models (LLMs) to human preferences.
RLHF contains three steps, i.e., human preference collecting, reward learning, and policy optimization, which are usually performed serially.
Despite its popularity, however, (fixed) reward models may suffer from inaccurate off-distribution, since policy optimization continuously shifts LLMs' data distribution.
Repeatedly collecting new preference data from the latest LLMs may alleviate this issue, which unfortunately makes the resulting system more complicated and difficult to optimize.
In this paper, we propose reward learning on policy (RLP), an unsupervised framework that refines a reward model using policy samples to keep it on-distribution.
Specifically, an unsupervised multi-view learning method is introduced to learn robust representations of policy samples.
Meanwhile, a synthetic preference generation approach is developed to simulate high-quality preference data with policy outputs.
Extensive experiments on three benchmark datasets show that RLP consistently outperforms the state-of-the-art.
Our code is available at \url{https://github.com/AlibabaResearch/DAMO-ConvAI/tree/main/rlp}.

\end{abstract}

\section{Introduction}

Large language models (LLMs)~\citep{brown2020language,bommasani2021opportunities} have shown great promise in following open-ended user instructions~\citep{askell2021general,ouyang2022training,longpre2023flan}.
These capabilities are largely attributed to the fine-tuning of pretrained LLMs using Reinforcement Learning from Human Feedback (RLHF)~\citep{christiano2017deep,bai2022training}, which is a prominent technique to align LLMs with human preferences and greatly enhance their usability and safety~\citep{OpenAI2023Gpt-4,Anthropic2023Introducing,Google2023Bard}.

\begin{figure}[t]
\centering
\includegraphics[width = 1.0\linewidth]{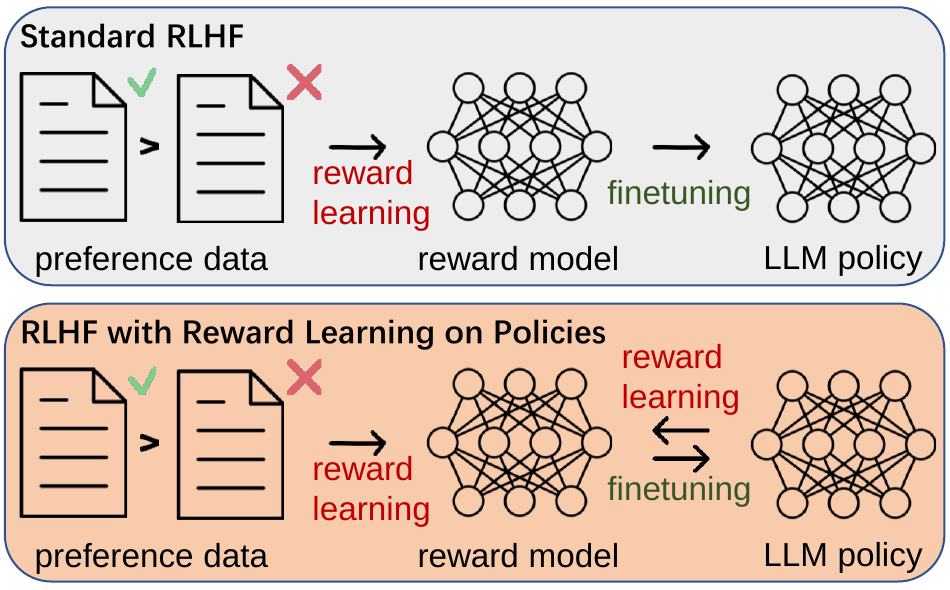}
\caption{Comparison of standard RLHF (top) and RLHF with reward learning on policies (bottom). Different from (top), which performs reward learning and policy optimization serially, we iteratively train one of the two models with the help of the other.}
\label{fig:idea}
\end{figure}

A typical RLHF procedure is comprised of three interrelated steps: human preference  collecting, reward learning, and policy optimization (Figure~\ref{fig:idea} top).
The reward learning step fits a reward model to the preference data that elicits evaluations from humans.
The policy optimization step uses reinforcement learning (RL) to fine-tune a language model to produce outputs assigned high reward.

In practice, the three key steps of RLHF are often performed serially~\citep{casper2023open}.
Since policy optimization shifts the language model’s data distribution during the RL phase, the (fixed) reward model will be inaccurate off-distribution which is trained on offline data~\citep{touvron2023llama}.
Hence, reward model accuracy can quickly degrade and in turn degenerate the policy that exploits differences between the inferred and true reward~\citep{gao2023scaling}.

The above issue can be mitigated by gathering new human preference data from an up-to-date version of policy~\citep{ziegler2019fine}.
However, the resulting system is significantly more complicated and difficult to optimize, involving iterations of data gathering, reward learning, and RL fine-tuning.
Moreover, significant work is required to maintain high data quality over a long time in this setting.

In this paper, we show how to optimize a reward model against the policy to keep it on-distribution, without repeatedly collecting new human preference data.
We propose \textit{Reward Learning on Policy (RLP)}, a framework that refines a reward model using policy samples in an unsupervised manner.
RLP first trains a reward model and a language model policy from scratch with standard RLHF methods, and then retrains the reward model when exposed to the sample distribution of the trained policy.
Finally, RLP retrains the policy on the retrained reward model, which attempts to maintain an accurate reward for the latest policy.

Concretely, RLP uses policy samples to retrain the reward model via two methods: unsupervised multi-view learning (UML) and synthetic preference generation (SPG).
RLP-UML constructs two views for an input by generating two responses from the policy~\citep{zhao2017multi}, then optimizes a multi-view information bottleneck loss~\citep{federici2020learning} when fitting the reward model to a dataset of human preferences.
This training objective follows the information bottleneck principle~\citep{tishby2000information} and helps learn robust representations of the policy's data distribution.

In addition, RLP-SPG simulates preferences on policy generations to supplement the human preference data.
Rather than producing and scoring two outputs with LLMs as in Reinforcement Learning from AI Feedback (RLAIF)~\citep{bai2022constitutional,lee2023rlaif}, RLP-SPG generates a set of outputs for an instruction.
In this way, RLP-SPG can quantify uncertainty and decide when to trust model predictions via measuring the size of the largest semantic equivalence cluster~\citep{kuhn2023semantic,si2022prompting}.
Thus, RLP-SPG selectively generates pairwise preferences for instructions with low uncertainty~\citep{lin2023generating}, where the preferred output is sampled from the largest cluster of the output set and the non-preferred one is sampled from the rest clusters.
This sampling scheme also conforms to the self-consistency assumption, i.e., the most consistent output is selected as the final prediction~\citep{wang2022self,chen2023teaching}.

Our main contributions are as follows:

\begin{itemize}
    \item We propose Reward Learning on Policy (RLP), an unsupervised framework that refines a reward model using policy samples to keep it on-distribution for RLHF. 
    \item We optimize a multi-view loss when retraining the reward model to learn representations of the policy's data distribution. We also simulate preferences with a set of policy outputs, which enables selective generation and high-quality data construction.
    \item Our experiments on three standard benchmark datasets show that RLP outperforms existing methods for learning from human feedback, including PPO-based RLHF.
\end{itemize}

\section{Related Work}

\textbf{Instruction tuning} is a procedure to fine-tune pre-trained LLMs with instructions and human-written completions~\citep{mishra2021cross,sanh2021multitask}, which increases the usability of LLMs~\citep{chung2022scaling}.
Recently, \textbf{RLHF} has emerged as the central method for fine-tuning LLMs based on human preferences and further improves their downstream task performance and alignment with user intent~\citep{christiano2017deep}.
Generally, RLHF methods first fit a reward model to human preferences, then fine-tune a language model to maximize the inferred reward using RL algorithms.

\textbf{Reward models} tend to be an imperfect estimate of the true reward due to misspecification~\citep{biyik2022learning} and misgeneralization~\citep{tien2022causal}, and imperfect in reward models leads to reward hacking~\citep{skalse2022defining}.
Methods with reward ensemble~\citep{coste2023reward} and diverse feedback~\citep{yu2023constructive} are proposed to tackle this issue.
Our method retrains the reward model with policy samples to make it on-distribution and generalize to the policy’s data distribution.

\textbf{Human feedback simulation} aims to generate additional synthetic preference data using weak human supervision and LLMs~\citep{bai2022constitutional}.
RLAIF approaches obtain pairwise preferences by scoring two outputs from a shared prompt~\citep{lee2023rlaif}, whereas RLCD generates outputs from two variants  of a prompt~\citep{yang2023rlcd}.
Our method RLP-SPG is the first attempt to simulate human preferences using a set of outputs.

\textbf{Uncertainty quantification} provides confidence scores for generations of LLMs, helping users decide when to trust these generation results~\citep{si2022prompting}.
Supervised methods fine-tune the language model to predict the uncertainty~\citep{kadavath2022language,lin2022teaching}, while unsupervised methods measure uncertainty by calculating semantic entropy or semantic dispersion amongst generated answers~\citep{kuhn2023semantic,lin2023generating}.
In this work, we measure uncertainty to selectively generate preference data.

\section{Preliminaries}

We start by introducing the instruction following task~\citep{ouyang2022training,bai2022training}.
Given user instructions $x \in \mathcal{X}$ (e.g., ``Generate a definition for artificial intelligence''), we aim to develop a model $\pi_{\theta}$ that generates high-quality responses $y \sim \pi_{\theta}(y|x)$ as judged by some latent reward model.
In this study, we focus on RLHF for this task, due to its central role in instruction-following LLMs~\citep{ouyang2022training}.
RLHF usually consists of three steps: human preference collecting, reward modeling, and RL policy optimization~\citep{dubois2023alpacafarm,rafailov2023direct,casper2023open}.

\paragraph{Step 0, SFT:} RLHF generally begins with a pre-trained model, which is fine-tuned with supervised learning on instruction-following demonstrations $(x, y)$, to produce a model $\pi^{\mathrm{SFT}}(y|x)$.

\paragraph{Step 1, Human preference collecting:} The first step is to produce pairs of responses $(y_1, y_2) \sim \pi^{\mathrm{SFT}}(y|x)$ for the instruction $x$.
These are then presented to humans who express preferences for each response, denoted as $y_w \succ y_l\ |\ x$ where $y_w$ and $y_l$ denotes the preferred and non-preferred completion amongst $(y_1, y_2)$ respectively.

\paragraph{Step 2, Reward learning:} The second step is to fit a reward model $r_\phi(x,y)$ by minimizing the negative log-likelihood loss~\citep{christiano2017deep}:
\begin{small}
\begin{equation} 
\mathcal{L}_{R}=-\mathbb{E}_{(x,y_w,y_l) \sim \mathcal{D}}[\log \sigma (r_\phi(x,y_w)-r_\phi(x,y_l))], \nonumber
\end{equation}
\end{small}

\noindent where $\mathcal{D}=\{(x,y_w,y_l)\}$ is a dataset of pairwise preferences and $\sigma$ is the sigmoid function. $r_\phi(x,y)$ is often initialized from $\pi^{\mathrm{SFT}}(y|x)$ with one additional linear layer that infers the reward value.

\paragraph{Step 3, RL policy optimization:} The third step is to use the reward model $r_\phi(x,y)$ to fine-tune the language model.
The parameters $\theta$ of $\pi$ are trained to maximize
\begin{small}
\begin{equation}
\mathbb{E}_{x \sim \mathcal{U},y \sim \pi_{\theta}(y|x)}[r_\phi(x,y)-\beta \mathbb{D}_{\mathrm{KL}}(\pi_{\theta}(y|x)||\pi_{\mathrm{ref}}(y|x))], \nonumber
\end{equation}
\end{small}

\noindent where $\mathcal{U}=\{x\}$ is an unlabeled instruction dataset, the language model policy $\pi_{\theta}(y|x)$ is fine-tuned from the SFT model $\pi^{\mathrm{SFT}}$ , the reference policy $\pi_{\mathrm{ref}}$ is also the SFT model $\pi^{\mathrm{SFT}}$, and $\beta$ is a regularization coefficient controlling the deviation from $\pi_{\mathrm{ref}}$.
This objective is typically optimized with RL algorithms such as PPO~\citep{schulman2017proximal}.

\section{Reward Learning on Policy}

\subsection{Overview}

In this study, we propose a novel RLHF framework to fine-tune LLMs with human feedback following five steps:
\textbf{Step 1-3.} Collect a pairwise human preference dataset $\mathcal{D}$, then train a reward model $r_\phi$ and fine-tune a language model policy $\pi_{\theta}$;
\textbf{Step 4.} Retrain a reward model $\hat{r}_\phi$ using outputs of policy $\pi_{\theta}$;
\textbf{Step 5.} Retrain a policy $\hat{\pi}_{\theta}$ based on the retrained reward model $\hat{r}_\phi$.

Before applying RLP, we assume existing RLHF approaches can be used to train the reward model $r_\phi$ and the policy $\pi_{\theta}$~\citep{ouyang2022training}.
The sample distribution of the policy $\pi_{\theta}$ can be quite different from the preference data $\mathcal{D}$ on which the reward model $r_\phi$ is trained~\citep{touvron2023llama}.
For example, outputs become increasingly longer after applying RLHF methods as shown in the analysis of AlpacaFarm~\citep{dubois2023alpacafarm}.
The average length of SFT outputs is 278 characters and applying PPO increases it to 637 tokens.
These distributional differences make the reward model $r_\phi$ inaccurate off-distribution.

Our goal is to refine the reward model using samples of the policy $\pi_{\theta}$ and keep it on-distribution.
This process is expected to increase the generalization of the retrained reward model $\hat{r}_\phi$ to policy samples.
Accordingly, it can maintain an accurate reward during the RL policy optimization phase.

\subsection{Reward Retraining}

We now describe how to retrain the reward model $\hat{r}_\phi$ in \textbf{Step 4} of RLP.
We first construct a dataset of policy samples $\mathcal{P}=\{(x, \boldsymbol{y})\ |\ x\in \mathcal{U}, \boldsymbol{y} \sim \pi_{\theta}(y|x)\}$, where $\boldsymbol{y}$ is a set of $n$ outputs from  policy $\pi_{\theta}$ for instruction $x$.
Then, we refine the reward model with policy samples $\mathcal{P}$ in addition to the human preference dataset $\mathcal{D}$.
Specifically, we propose two different methods for this purpose: unsupervised multi-view learning (UML)  and synthetic preference generation (SPG).

\paragraph{Unsupervised Multi-View Learning} attempts to learn robust representations of policy samples.
For each pair $(x, \boldsymbol{y}) \in \mathcal{P}$, two semantic invariant views are constructed: $v_i(x)=(x, y)\ |\ y \sim \boldsymbol{y}$, $(i=1,2)$.
These two views preserve the same task-relevant information \citep{zhao2017multi}.
Then, a multi-view information bottleneck (MIB) loss~\citep{federici2020learning} is optimised for unsupervised representation learning, following the information bottleneck principle \citep{tishby2000information}.
This optimization process retains task-relevant information in the representations while discarding superficial information.

To facilitate the computation, we parametrize the representation $z_i$ of each view $v_i(x)$ with a factorized Gaussian distribution, i.e., $p_{\psi}(z|v_i) = \mathcal{N}[\mu(v_i),\Sigma(v_i)]$.
Concretely, we estimate $v_i(x)$ with the final transformer layer of the reward model and use two neural networks $\mu(v_i)$ and $\Sigma(v_i)$ to produce the mean and deviation respectively.
The following MIB loss is optimized:

\begin{small}
\begin{equation} 
\mathcal{L}_{M}=\mathbb{E}_{(x, \boldsymbol{y}) \sim \mathcal{P}}[-\mathbb{I}(z_1;z_2)+\mathbb{D}_{\mathrm{SKL}}(p_\psi(z|v_1)||p_\psi(z|v_2))],\nonumber
\end{equation}
\end{small}

\noindent where $\mathbb{I}$ calculates mutual information of two random variables, and $\mathbb{D}_{\mathrm{SKL}}$ represents the symmetrized KL divergence obtained by averaging the expected value
of $\mathbb{D}_{\mathrm{KL}}(p_\psi(z|v_1)||p_\psi(z|v_2))$ and $\mathbb{D}_{\mathrm{KL}}(p_\psi(z|v_2)||p_\psi(z|v_1))$.

\paragraph{Synthetic Preference Generation} aims to simulate high-quality preference data with policy samples.
For each pair $(x, \boldsymbol{y}) \in \mathcal{P}$, we assume the most frequent item of $\boldsymbol{y}$ as the correct prediction and its frequency as the confidence score~\citep{si2022prompting}, following the self-consistency assumption~\citep{wang2022self}.
To address semantic equivalence, i.e., different sentences can mean the same thing, we cluster items of $\boldsymbol{y}$ into groups $\mathcal{G}$ with a bi-directional entailment algorithm~\citep{kuhn2023semantic}.
Sentences from each group $\boldsymbol{g} \in \mathcal{G}$ are expected to share the same meaning.
We estimate the confidence score of $(x, \boldsymbol{y})$ as $\frac{|\tilde{\boldsymbol{g}}|}{|\boldsymbol{y}|}$, where $\tilde{\boldsymbol{g}}$ is the largest group of $\mathcal{G}$ and the operator $|\cdot|$ measures the size of a set.
Thus, we can selectively generate a synthetic preference dataset with high confidences $\hat{\mathcal{D}}=\{(x,y_w,y_l)\ |\ (x, \boldsymbol{y}) \in \mathcal{P},\frac{|\tilde{\boldsymbol{g}}|}{|\boldsymbol{y}|}\ge \gamma,y_w \sim \tilde{\boldsymbol{g}},  y_l \sim \boldsymbol{y}\setminus \tilde{\boldsymbol{g}}\}$, where $\gamma$ is the threshold for selective generation, the preferred output $y_w$ is sampled from the largest group $\tilde{\boldsymbol{g}}$ with the largest reward score and the non-preferred one $y_l$ is randomly sampled from the rest groups.

The overall loss that we optimize for the reward model $\hat{r}_\phi$ is:

\begin{small}
\begin{equation} 
\begin{aligned}
\mathcal{L}_{\hat{R}}=&-\mathbb{E}_{(x,y_w,y_l) \sim \mathcal{D} \cup \hat{\mathcal{D}}}[\log \sigma (\hat{r}_\phi(x,y_w)-\hat{r}_\phi(x,y_l))] \\
                       &+ \lambda\ \mathcal{L}_{M},
\end{aligned}
\label{eq:reward}
\end{equation}
\end{small}

\noindent where the coefficient $\lambda$ controls the weight of the multi-view information bottleneck loss.
To simplify computational complexity, we implement two variants:
1. \textbf{RLP-UML} removes the synthetic dataset $\hat{\mathcal{D}}$ in Eq.~\ref{eq:reward} and learns the representations of policy samples when fitting the reward model.
2. \textbf{RLP-SPG} removes the MIB loss by setting $\lambda=0$ in Eq.~\ref{eq:reward} and fits the reward model with human and synthetic preference data.

\subsection{Policy Retraining}

We finally retrain the policy $\hat{\pi}_{\theta}$ using $\hat{r}_\phi$ in \textbf{Step 5} of RLP. 
Specifically, we optimize $\hat{\pi}_{\theta}$ to maximize

\begin{small}
\begin{equation}
\mathbb{E}_{x \sim \mathcal{U},y \sim \hat{\pi}_{\theta}(y|x)}[\hat{r}_\phi(x,y)-\beta \mathbb{D}_{\mathrm{KL}}(\hat{\pi}_{\theta}(y|x)||\pi_{\mathrm{ref}}(y|x))]. \nonumber
\end{equation}
\end{small}

Our approach RLP is summarized in Algorithm~\ref{alg:algorithm}.

\begin{algorithm}
\caption{RLP: RLHF with Reward Learning on Policy}
\label{alg:algorithm}
\KwIn{SFT model $\pi^{\mathrm{SFT}}$, unlabeled data $\mathcal{U}$.}
\KwOut{A language model policy $\hat{\pi}_{\theta}$.}
Collect a human preference dataset $\mathcal{D}$.\\
Train a reward model $r_\phi$ using $\mathcal{D}$.\\
Fine-tune a language model $\pi_{\theta}$ from $\pi^{\mathrm{SFT}}$ using $\mathcal{U}$ and $r_\phi$.\\
Retrain a reward model $\hat{r}_\phi$ using $\mathcal{L}_{\hat{R}}$ (Eq.~\ref{eq:reward}). \\
Fine-tune $\hat{\pi}_{\theta}$ from $\pi^{\mathrm{SFT}}$ using $\mathcal{U}$ and $\hat{r}_\phi$.\\
\end{algorithm}

\begin{table}[htp]
    \centering
    \small
    \begin{tabular}{l | l |c}
    \toprule
   \multicolumn{2}{c|}{Dataset} & \#Sample \\
    \midrule
    \multirow{4}{*}{Training data} & SFT dataset & 10k \\
                  \cmidrule{2-3}  
                  & Preference dataset $\mathcal{D}$ & 10k \\
                  \cmidrule{2-3} 
                  & Unlabeled dataset $\mathcal{U}$ & 20k \\
    \midrule
    \multirow{4}{*}{Evaluation data} & AlpacaFarm & 805 \\
                  \cmidrule{2-3}  
                  & LLMBar & 100 \\
                  \cmidrule{2-3} 
                  & Vicuna & 80 \\
    \bottomrule
    \end{tabular}
    \caption{Dataset statistics.}
    \label{tab:data}
\end{table}

\section{Experiments}

\begin{table*}[htb]
    \centering
    \small
    \setlength\tabcolsep{4pt} 
    \begin{tabular}{l|cc|c|c}
    \toprule
    \multicolumn{1}{l|}{\multirow{2}{*}{\textbf{Method}}} & \multicolumn{2}{c|}{\textbf{AlpacaFarm}}  & \textbf{LLMBar} & \textbf{Vicuna} \\
    \multicolumn{1}{l|}{} & Simulated Win-Rate & Human Win-Rate &  Simulated Win-Rate & Simulated Win-Rate \\
    \midrule
    GPT-4 & 79.0 & 69.8 & 74.0 & 85.0 \\
    ChatGPT & 61.4 & 52.9 & 59.0 & 63.7 \\
    PPO & 46.8 & 55.1 & 47.5 & 57.5 \\
    Best-of-$n$ & 45.0   & 50.7 & 43.4 & 52.5 \\
    SFT & 36.7 & 44.3 & 42.4 & 50.0 \\
    LLaMA-7B & 11.3 & 6.5 & 12.5 & 12.8 \\
    \midrule
    RLP-UML (ours) & 49.1 & 56.5 & 48.5 & 61.3 \\
    \textbf{RLP-SPG} (ours) & \textbf{50.2} & \textbf{57.4} & \textbf{50.5} & \textbf{62.5} \\
    \bottomrule
    \end{tabular}
    \caption{The win-rate (\%) performance of RLP and baselines. Win-rates  are computed against reference model \texttt{text-davinci-003}. Baseline results in AlpacaFarm come from~\citet{dubois2023alpacafarm}. \textbf{Bold} numbers are superior results among the implemented LLMs. We omitted LLMBar and Vicuna for human evaluation because the simulated method rankings consistently correlate with the human method rankings in AlpacaFarm.}
    \label{tab:main_result}
\end{table*}

\begin{figure*}[htp]
\centering
\subfigure[]{
\begin{minipage}[t]{0.25\linewidth}
\centering
\label{fig:dis1}
\includegraphics[width=1.0\linewidth]{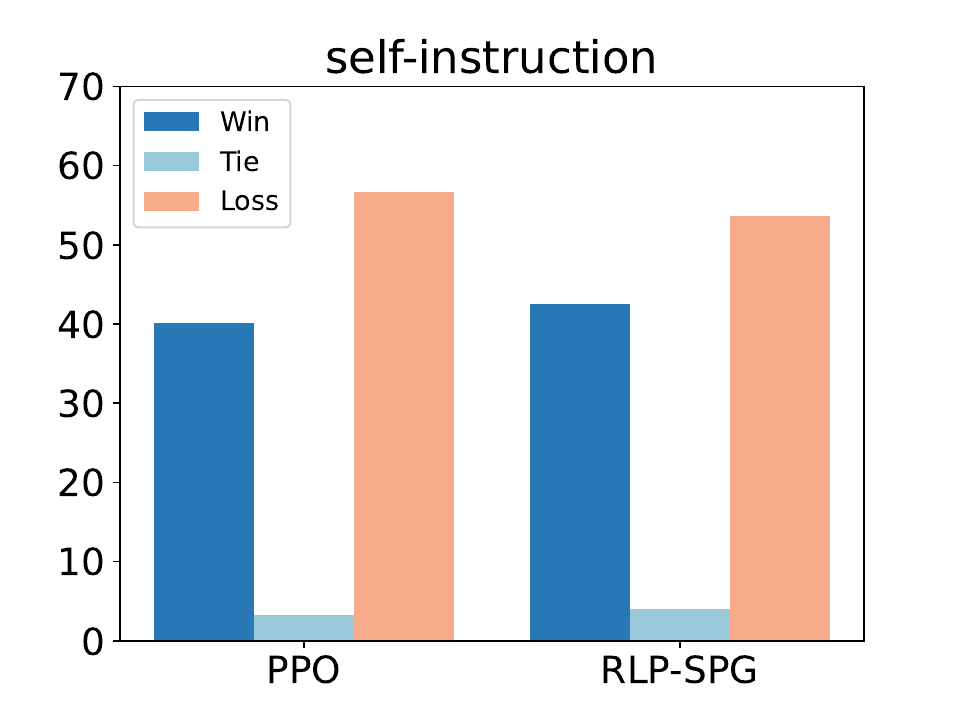}
\end{minipage}
}%
\subfigure[]{
\begin{minipage}[t]{0.25\linewidth}
\centering
\label{fig:dis2}
\includegraphics[width=1.0\linewidth]{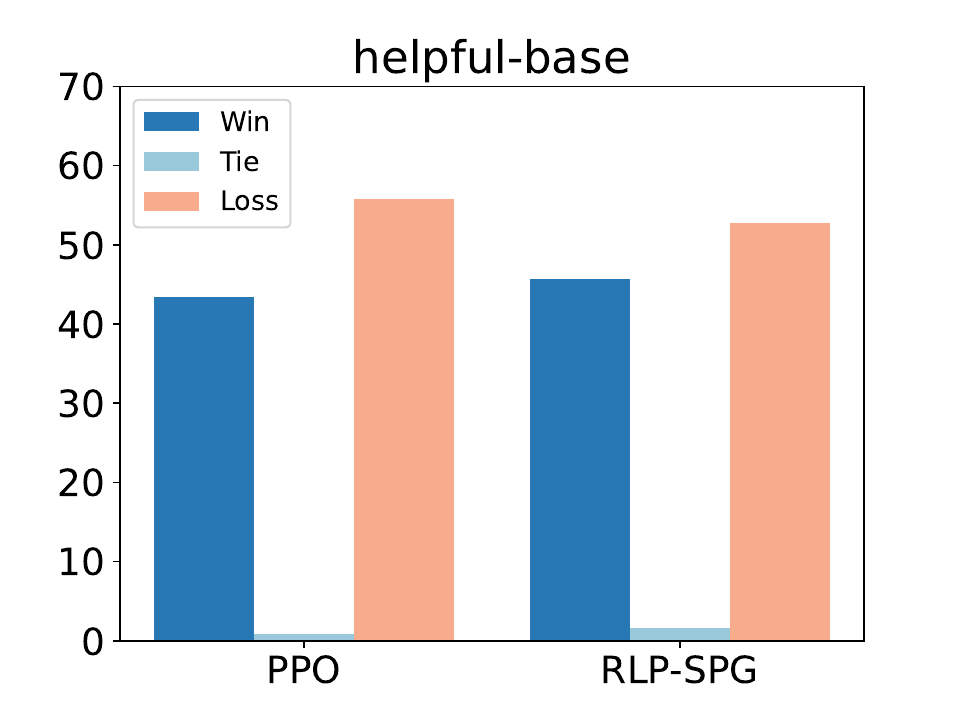}
\end{minipage}
}%
\subfigure[]{
\begin{minipage}[t]{0.25\linewidth}
\centering
\label{fig:dis3}
\includegraphics[width=1.0\linewidth]{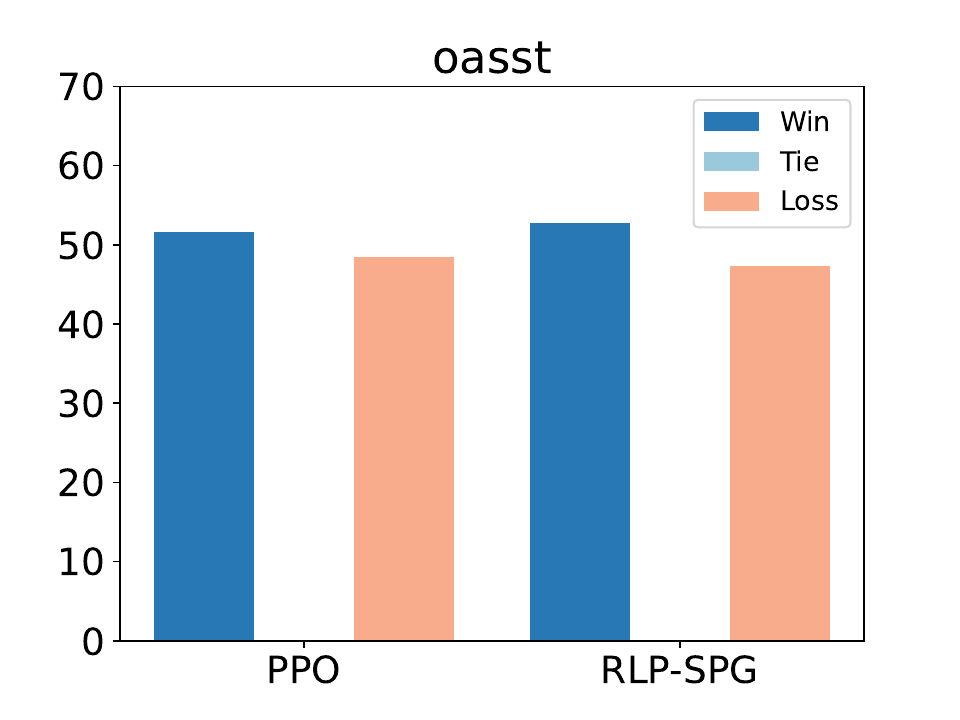}
\end{minipage}
}%
\subfigure[]{
\begin{minipage}[t]{0.25\linewidth}
\centering
\label{fig:dis3}
\includegraphics[width=1.0\linewidth]{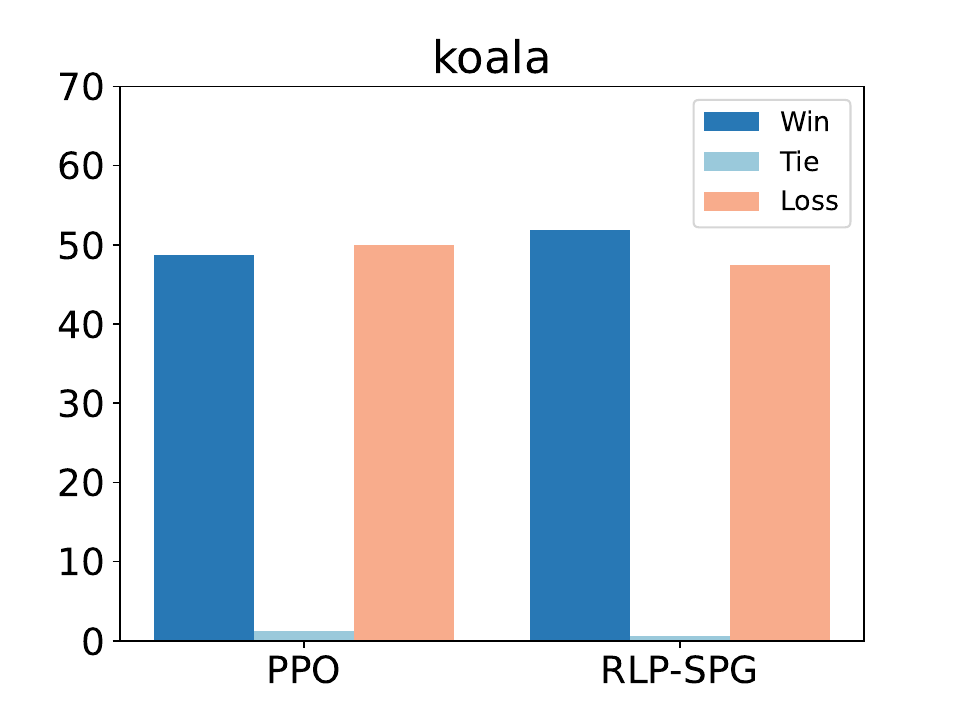}
\end{minipage}
}%
\centering
\caption{The simulated win-rate (\%) performance of RLP-SPG compared to PPO on various subsets of AlpacaFarm. Win-rates  are computed against reference model \texttt{text-davinci-003}.}
\label{fig:distribution}
\end{figure*}

\subsection{Datasets}

We run experiments on the instruction following task~\citep{ouyang2022training,bai2022training}, which remains a challenging task for the strongest LLMs today~\citep{wu2023reasoning,li2023instruction}.

\paragraph{Training data} of the RLHF procedure come from Alpaca data, which consists of 52k instruction-following demonstrations $(x, y)$~\citep{taori2023stanford}.
Following the data splits of AlpacaFarm~\citep{dubois2023alpacafarm}, three splits are used:
1. SFT split are 10k data for fine-tuning the SFT model $\pi^{\mathrm{SFT}}$;
2. Preference split are 10k instructions on which we collect pairwise preference dataset $\mathcal{D}$;
3. Unlabeled split are 20k unlabeled instructions $\mathcal{U}$ used in PPO.
Concretely, two variants of preference dataset $\mathcal{D}$ are curated: simulated $\mathcal{D}_{\mathrm{sim}}$ are constructed with AlpacaFarm simulated annotators by prompting API LLMs, and human $\mathcal{D}_{\mathrm{human}}$ are constructed with human annotators.

\paragraph{Evaluation data} of the trained LLMs include three benchmarks:
1. \textbf{AlpacaFarm} consists of 805 instructions, which includes instructions from Self-instruct evaluation~\citep{wang2022super}, OASST evaluation, Anthropic helpful evaluation~\citep{bai2022training}, Vicuna evaluation~\citep{chiang2023can}, and Koala evaluation~\citep{geng2023koala};
2. \textbf{LLMBar}~\citep{zeng2023evaluating} contains 100 instructions with objective preferences. We use the natural set to evaluate performance in real-world distributions;
3. \textbf{Vicuna}~\citep{chiang2023vicuna} contains 80 instructions to evaluate different skills of LLMs.
Table~\ref{tab:data} shows the statistics of these datasets.

\begin{table*}[t]
    \centering
    \small
    \begin{tabular}{p{12cm} }
    \toprule
    \textbf{Instruction:}\\
    What are different drawers I should have for clothes? \\
    \midrule
    \textbf{SFT Output:}\\
    Different drawers for clothes can include a top drawer for socks and underwear, a middle drawer for shorts, pants, and skirts, a lower drawer for long-sleeved shirts, a separate drawer for t-shirts and tank tops, a drawer for sweaters and cardigans, a drawer for jackets, and a drawer for shoes. \\
    \midrule
    \textbf{PPO Output:}\\
    Different drawers for clothes can include a master drawer for all items, such as pants, shirts, and jackets; a drawer for socks and underwear; a drawer for delicates, such as lingerie and fine clothing; a drawer for seasonal clothing, such as winter coats and swimwear; a drawer for work clothes, such as dress shirts and skirts; a drawer for shoes; and a drawer for accessories, such as belts, hats, and purses.\\
    \midrule
    \textbf{RLP-UML Output:}\\
    Different drawers for clothes can include:\\
    - A top drawer for long-sleeved shirts and blouses.\\
    - A middle drawer for short-sleeved shirts and blouses.\\
    - A bottom drawer for sweaters, sweaters, and jackets.\\
    - A separate drawer for pants, shorts, and skirts.\\
    - A drawer for socks and underwear.\\
    - A drawer for accessories such as belts, scarves, and hats.\\
    - A separate drawer for seasonal clothing, such as winter coats and summer dresses.\\
    \midrule
    \textbf{RLP-SPG Output:}\\
    Different drawers for clothes can include:\\
    - A master drawer for folded clothing, such as shirts, pants, skirts, and shorts. \\
    - A drawer for seasonal items, like sweaters and jackets.\\
    - A drawer for delicate items, like lingerie and fine fabrics.\\
    - A drawer for formal wear, like suits and dresses.\\
    - A drawer for accessories, like belts, scarves, and jewelry.\\
    - A drawer for uniforms, like work clothes and school uniforms.\\
    - A drawer for out-of-season items, like swimwear and winter coats.\\
    - A drawer for shoes, organized by type and use.\\
    - A drawer for cleaning supplies, like a laundry bin or ironing board.\\
    - A drawer for miscellaneous items, like first-aid kits and sewing supplies.\\
    \bottomrule
    \end{tabular}
    \caption{Outputs generated by RLP and baselines for an example from AlpacaFarm. RLP-UML and RLP-SPG produce more comprehensive outputs than SFT and PPO in this case.}
    \label{tab:case}
\end{table*}

\subsection{Metrics and Experimental Setups}

Following~\citet{dubois2023alpacafarm,touvron2023llama}, we use win-rate to evaluate the performance of an LLM $\pi_{\theta}$, i.e., the percentage of times $\pi_{\theta}$ is preferred to a reference model $\pi_{\mathrm{ref}}$ for their instruction-following outputs.
Following the experimental setups of AlpacaFarm~\citep{dubois2023alpacafarm}, we use simulated win-rate to evaluate methods trained on simulated $\mathcal{D}_{\mathrm{sim}}$ with simulated annotators by prompting API LLMs.
In parallel, we use human win-rate to evaluate methods trained on human $\mathcal{D}_{\mathrm{human}}$ with human annotators.

\subsection{Implementation Details}

Our implementations of reward learning, policy optimization, and simulated annotators are based on the AlpacaFarm codebase and its default hyperparameters~\citep{dubois2023alpacafarm}.
Specifically, we use RL algorithm PPO~\citep{schulman2017proximal} to fine-tune the language model during policy optimization.
All reward models and language models are based on LLaMA-7B~\citep{touvron2023llama1}.
We sample $n=10$ outputs for each instruction to construct a dataset of policy samples $\mathcal{P}$.
For unsupervised multi-view learning, we implement $\mu(v_i)$ and $\Sigma(v_i)$ as three-layer MLPs, and use Jensen-Shannon mutual information estimator \citep{hjelm2018learning} to estimate mutual information $\mathbb{I}$ in the MIB loss.
For synthetic preference generation, we implement a bidirectional entailment clustering algorithm using Deberta-large model~\citep{he2020deberta} and set the threshold $\gamma=0.5$ for selective generation.
We set $\lambda=0.5$ in Eq.~\ref{eq:reward} for RLP-UML.
At training and inference time, we use a sampling temperature of 1.0 and 0.7, respectively.
All experiments are performed on a single 8$\times$A100 machine.

\subsection{Baselines}

We compare RLP with competitive baselines:
\textbf{1. LLaMA-7B}~\citep{touvron2023llama1} directly generates outputs using the base unaligned LLaMA-7B;
\textbf{2. SFT}~\citep{taori2023stanford} is a LLaMA-7B model supervised fine-tuned on 10k Alpaca instruction-following data;
\textbf{3. Best-of-$n$}~\citep{stiennon2020learning} samples $n$ i.i.d. responses from the SFT model and returns the response with the highest inferred reward;
\textbf{4. PPO}~\citep{schulman2017proximal} is a reinforcement learning algorithm that maximizes surrogate reward, subject to a KL penalty keeping parameters near the SFT model;
\textbf{5. ChatGPT} uses OpenAI API LLM \texttt{gpt-3.5-turbo-0301};
\textbf{6. GPT-4} uses OpenAI API LLM \texttt{gpt-4-0314}.

\subsection{Main Results}

We compare the win-rate performance of our method RLP and all baselines on three standard benchmarks to assess their instruction-following ability in Table~\ref{tab:main_result}.
It can be seen that API LLM GPT-4 significantly outperforms all other models due to its obvious advantages.
Among the implemented LLMs, RLP-SPG performs the best in both the simulator and human preference data, and achieves SOTA results on all three benchmarks.
Compared with the implemented best-performing baseline PPO, RLP-SPG brings up from a simulator win-rate of 46.8\% to 50.2\% in AlpacaFarm, 47.5\% to 50.5\% in LLMBar, and 57.5\% to 62.5\% in Vicuna.
RLP-SPG also brings up from a human win-rate of 55.1\% to 57.4\% in AlpacaFarm.

We can also observe that:
\textbf{1.} Both the two variants of RLP, namely, RLP-UML and RLP-SPG, outperform all implemented baselines that do not train reward models using policy samples.
The performance gain demonstrates the advantage of considering policy for reward learning, which can help keep the reward model on-distribution.
\textbf{2.} RLP-SPG generally outperforms RLP-UML under all circumstances.
It demonstrates that synthetic preference generation leads to better performance, which simulates pairwise preference data with policy samples that can be leveraged for optimizing a reward model directly.

To provide a clearer perspective on RLP’s superiority over other baselines, we illustrate the simulated win-rate of our best method RLP-SPG compared to the best-performing baseline PPO on various subsets of AlpacaFarm in Figure~\ref{fig:distribution}.
Instructions from these subsets show diverse coverage over realistic interactions, allowing for an intricate analysis of the proficiency attained through language model fine-tuning~\citep{dubois2023alpacafarm}.
Notably, RLP-SPG outperforms PPO across all subsets, including Self-instruct evaluation, Anthropic helpful evaluation, OASST evaluation, and Koala evaluation.
It further indicates that reward learning on policy leads to a comprehensive enhancement in the capabilities of the LLMs.
Meanwhile, RLP also outperforms these baselines on knowledge intensive benchmarks such as MMLU~\citep{hendrycks2020measuring} (See Appendix \ref{append:eval_knowledge}.).

The difference between RLP and baselines can be observed qualitatively as well.
For example, the case shown in Table~\ref{tab:case} makes it sufficiently clear why RLP is so strongly preferred over our baselines from AlpacaFarm.
Compared to RLP-UML, RLP-SPG generates even longer and more comprehensive outputs.

\subsection{Ablation Studies}

This section provides comprehensive ablation studies to understand the efficacy of RLP.
For consistency, all ablations are conducted using metric simulated win-rate that is computed against reference model \texttt{text-davinci-003}.

\begin{table}[htp]
    \centering
    \small
    \begin{tabular}{l|c|c}
    \toprule
    \multirow{2}{*}{\textbf{Method}} & \textbf{AlpacaFarm} & \textbf{LLMBar} \\
     &  Win-Rate & Win-Rate \\
    \midrule
    InfoMax & 44.4 & 46.5 \\
    MVI & 48.1 & 47.5 \\
    CL & 48.2 & 46.5 \\
    \midrule
    \textbf{RLP-UML} & \textbf{49.1} & \textbf{48.5} \\
    \bottomrule
    \end{tabular}
    \caption{Ablation study on the representation learning loss for RLP-UML.}
    \label{tab:ab_rep}
\end{table}

\paragraph{Information Bottleneck Loss}
We demonstrate the effectiveness of our multi-view information bottleneck loss $\mathcal{L}_{M}$ by replacing $\mathcal{L}_{M}$ in Eq.~\ref{eq:reward} with other alternatives of representation learning:
\textbf{1. InfoMax} \citep{poole2019variational} maximizes the mutual information between an input $v(x)$ and its representation $z$, i.e., $\mathbb{I}(v;z)$;
\textbf{2. MVI} \citep{bachman2019learning} is similar to InfoMax except that it maximizes the mutual information between its two views $\mathbb{I}(z_1;z_2)$;
Note that neither InfoMax nor MVI attempts to remove superficial information from representations.
\textbf{3. CL} \citep{caron2020unsupervised} uses a contrastive learning loss. Positive pairs in this variant are obtained using our multi-view construction approach.

Results in Table \ref{tab:ab_rep} show that the information bottleneck loss used in RLP-UML performs better than all other variants.
We also want to highlight that the approach of explicitly removing superficial information in RLP-UML makes it outperform InfoMax and MVI by $4.7\%$ and $1.0\%$ in AlpacaFarm, and $2.0\%$ and $1.0\%$ in LLMBar, respectively.
This validates our claim that the multi-view information bottleneck loss used in RLP-UML effectively removes the superficial information.

\begin{table}[t]
    \centering
    \small
    \begin{tabular}{l|l|c|c}
    \toprule
    \multirow{2}{*}{\textbf{Sampling}} & \multirow{2}{*}{\textbf{Method}} & \textbf{Preference} & \textbf{AlpacaFarm} \\
     & & Acc. & Win-Rate \\
    \midrule
    \multirow{2}{*}{A Pair} & RLAIF & 52 & 46.0 \\
    & Reward & 56 & 48.9 \\
    \midrule
    \multirow{3}{*}{A Set} & Select-All & 55 & 48.7 \\
    \cmidrule{2-4}
    & \textbf{RLP-SPG} & \textbf{59} & \textbf{50.2} \\
    \bottomrule
    \end{tabular}
    \caption{Ablation on different synthetic preference generation approaches.}
    \label{tab:ab_spg}
\end{table}

\begin{table*}[htp]
    \centering
    \small
    \begin{tabular}{p{12cm} }
    \toprule
    \textbf{Instruction:}\\
    Generate a definition for the term ``artificial intelligence''.\\
    \midrule
    \textbf{RLAIF Simulated Preference Data:} \\
    \textbf{$y_w$:} Artificial intelligence (AI) is a branch of computer science that enables machines to learn from experience, adjust to new input, and mimic human decision-making processes. AI technology is used to develop systems that can interact with their environment and achieve tasks such as robotic navigation, natural language processing, and computer vision. \\
    \textbf{$y_l$:} Artificial intelligence (AI) is a field of study focused on developing computers and machines that are capable of performing tasks that are typically done by humans, such as reasoning, problem solving, and decision-making. AI technology is used in a variety of applications, including robotics, games, speech recognition, natural language processing, and automation. \\
    \midrule
    \textbf{RLP Simulated Preference Data:} \\
    \textbf{$y_w$:} Artificial intelligence (AI) is a branch of computer science that involves the use of sophisticated algorithms and technology to enable machines to think, learn, and act like humans. AI systems can be programmed to perform a variety of tasks, from simple to complex, such as recognizing speech, understanding natural language, automating tasks, and more. AI technology is used in a wide range of applications, from self-driving cars to virtual assistants to automated customer service. In short, AI is the science and technology of making computers and machines smart and capable of doing things traditionally done by humans. \\
    \textbf{$y_l$:} Artificial intelligence (AI) is a field of study focused on developing computers and machines that can think and act like humans. AI technology allows machines to interact with their environment and to learn from their mistakes, just like humans do. \\
    \bottomrule
    \end{tabular}
    \caption{An example instruction and corresponding simulated preference data from RLAIF and RLP-SPG.}
    \label{tab:simulated_prompt}
\end{table*}

\begin{table*}[!h]
    \centering
    \small
    \begin{tabular}{l | l   c }
    \toprule
    \textbf{Decision} & \textbf{Instruction} & \textbf{Confidence} \\
    \midrule
    \multirow{4}{*}{Reject} & Describe the life and reign of King Charles II. & 0.1 \\
    & What type of fruit would be a good addition to a fruit salad? & 0.2 \\
    & Research about a famous scientist and provide a short biography about him/her. & 0.2 \\
    & Compose a five word sentence describing your day. & 0.4\\
    \midrule
    \multirow{4}{*}{Accept} & Write a scientific explanation for why the sky is blue. & 0.6 \\
    & Find the synonyms of the following word: `Tenacious'. & 0.7 \\
    & Find the main idea of the following passage. & 0.8 \\
    & Create a tweet summarizing the following news article in 140 characters or less. & 0.9 \\
    \bottomrule
    \end{tabular}
    \caption{Cases of rejected and accepted instructions for selective synthetic preference generation by RLP-SPG.}
    \label{tab:rej}
\end{table*}

\paragraph{Synthetic Preference Generation}
We compare RLP with two types of synthetic preference generation approaches:
\textbf{\uppercase\expandafter{\romannumeral1}.} sampling a pair of responses for each instruction and then labeling its preference with LLMs; 
\textbf{\uppercase\expandafter{\romannumeral2}.} sampling a set of outputs for each instruction and then selecting a preferred and a non-preferred one to construct pairwise preference data.
For type \textbf{\uppercase\expandafter{\romannumeral1}}, we implement \textbf{RLAIF}~\citep{lee2023rlaif} that labels preferences with policy $\pi_{\theta}$, and \textbf{Reward} that rank the two outputs with reward model $r_\phi$ and assume the top ranked output as the preferred one.
For type \textbf{\uppercase\expandafter{\romannumeral2}}, we study a variant of RLP, \textbf{Select-All}, that sets $\gamma=0$ for selective generation and no longer rejects low confidence data.

Table~\ref{tab:ab_spg} shows the accuracy of generated preference data and the win-rate of the corresponding LLMs.
Golden preferences are labeled with AlpacaFarm simulated annotator.
These results indicate that RLP-SPG outperforms all ablation variants in terms of synthetic preference quality and LLM performance.
We can also observe that:
1. Sampling a set of outputs rather than a pair for each instruction helps encourage output diversity and leads to high-quality preference generation.
2. The confidence score based on multiple sampling can be used for selective generation and further improve preference quality.
3. LMMs trained with more accurate preference data generally perform better and obtain higher win-rate scores.

\subsection{Further Analysis}

Here we present further analysis of intermediate results during LLM training.
Table~\ref{tab:simulated_prompt} shows an example of simulated preference data by RLAIF and RLP-SPG, respectively.
The two outputs ($y_w$ and $y_l$) of RLAIF for this case look quite similar. However, $y_w$ is preferred by RLAIF which would bring in noises in the training process.
On the contrary, $y_w$ of RLP-SPG is more comprehensive than $y_l$ of RLP-SPG, resulting in more accurate labels.
We also find a major difference in the length distributions of RLP-SPG outputs, with preferred outputs $y_w$ (510 characters on average) significantly longer than non-preferred outputs $y_l$ (449 characters on average).

Table~\ref{tab:rej} also demonstrates cases of rejected and accepted instructions for selective synthetic preference generation by RLP-SPG (rejecting low confidence generations).
It can be observed that open-ended instructions (e.g., more subjective and creative) tend to have low confidences and be rejected.

\section{Conclusion}
In this paper, we propose reward learning on policy (RLP), a novel framework to align LLMs with human preferences.
RLP learns robust representations of the policy's data distribution via optimizing a multi-view information bottleneck loss.
RLP also simulates preferences with a set of policy outputs, which enables confidence estimation and selective generation.
Extensive experiments demonstrate that RLP outperforms SOTA baselines.

\section*{Limitations}

While we have carefully studied the effectiveness of RLP compared to several baselines on three benchmark datasets for LLaMA-7B,  we have not yet empirically verified our conclusions when aligning larger pretrained LLMs.
It would also be interesting to align new SOTA pretrained LLMs such as LLaMA 2~\citep{touvron2023llama} and test other methods for fitting preference data like DPO~\citep{rafailov2023direct}.

Meanwhile, all of our training and evaluation data are in English, and we have not tested in other languages.
Performance may degenerate especially in low-resource languages when pretrained LLMs have not been trained with these data.

\section*{Ethics Statement}
This work does not raise any direct ethical issues.
In the proposed work, we seek to develop a novel RLHF framework to align large language models (LLMs) with human preferences.
Concretely, we propose to learn reward models on policy to keep it on-distribution.
We believe this work can benefit the field of LLMs, with the potential to benefit other
fields requiring NLP models.
All experiments are conducted on open datasets.

\bibliography{anthology,custom}

\appendix

\section{Evaluation on Knowledge Intensive Benchmark}\label{append:eval_knowledge}

We also evaluate the performance of RLP on knowledge intensive benchmark MMLU~\citep{hendrycks2020measuring}, which includes exam questions from 57 tasks such as mathematics, history, law, and medicine.
Specifically, we use InstructEval~\citep{chia2023instructeval} to perform the evaluation.
As shown in Table~\ref{tab:ki}, both RLP-UML and RLP-SPG outperform PPO.

\begin{table}[htp]
    \centering
    \small
    \begin{tabular}{l|c}
    \toprule
    \textbf{Method} & \textbf{MMLU} \\
    \midrule
    GPT-4 & 86.4 \\
    ChatGPT & 70.0 \\
    PPO & 36.9 \\
    LLaMA-7B & 35.2 \\
    \midrule
    \textbf{RLP-UML} (ours) & \textbf{37.6} \\
    RLP-SPG (ours) & 37.3 \\
    \bottomrule
    \end{tabular}
    \caption{Performance of RLP and baselines on the MMLU benchmark. The scores are obtained by running InstructEval.}
    \label{tab:ki}
\end{table}

\end{document}